\pdfoutput=1

\documentclass[11pt]{article}

\usepackage[]{EMNLP2023}
\usepackage{times}
\usepackage{latexsym}

\usepackage{adjustbox}
\usepackage{graphicx}
\usepackage{multirow}
\usepackage{amsmath}  
\usepackage{amsfonts} 
\usepackage{url}
\usepackage{subcaption}

\usepackage{arydshln}
\usepackage{booktabs} 

\usepackage[T1]{fontenc}

\usepackage[utf8]{inputenc}
\usepackage{microtype}
\usepackage{inconsolata}

\title{Instruction Position Matters in Sequence Generation \\ with Large Language Models}

\author{
  Yijin Liu,~
  Xianfeng Zeng,~ 
  Fandong Meng\thanks{\quad  Corresponding author.} ~
  and Jie Zhou \\
  Pattern Recognition Center, WeChat AI, Tencent Inc, China \\
  \texttt{\{yijinliu, xianfzeng, fandongmeng, withtomzhou\}@tencent.com} \\
}

\begin{document}
\maketitle

\begin{abstract}
Large language models (LLMs) are capable of performing conditional sequence generation tasks, such as translation or summarization, through instruction fine-tuning.
The fine-tuning data is generally sequentially concatenated from a specific task instruction, an input sentence, and the corresponding response.
Considering the locality modeled by the self-attention mechanism of LLMs, these models face the risk of \textit{instruction forgetting} when generating responses for long input sentences. 
To mitigate this issue, we propose enhancing the instruction-following capability of LLMs by 
shifting the position of task instructions after the input sentences. 
Theoretical analysis suggests that our straightforward method can alter the model's learning focus, thereby emphasizing the training of instruction-following capabilities. 
Concurrently, experimental results demonstrate that our approach consistently outperforms traditional settings across various model scales (1B / 7B / 13B) and different sequence generation tasks (translation and summarization), without any additional data or annotation costs. 
Notably, our method significantly improves the  zero-shot performance on conditional sequence generation, {\em e.g.,} up to 9.7 BLEU points on WMT zero-shot translation tasks.



\end{abstract}

\section{Introduction}

In recent years, there has been a rapid emergence of large language models (LLMs) like GPT-4 and ChatGPT\footnote{\url{https://chat.openai.com/chat}}, which have demonstrated excellent zero-shot capabilities without the need for supervised fine-tuning. These models have shown promising performance in various traditional natural language processing tasks~\cite{chatgpt1,chatgpt2,chatgpt3,chatgpt4,chatgpt5,chatgpt6}. However, there is also a growing interest in open-source medium-sized language models, such as the LLaMA model with 13 billion parameters~\cite{llama_2023} and the BLOOMZ language model with 7.1 billion parameters~\cite{bloomz_2022}, to meet the research and hardware deployment requirements.

To align the outputs of language models with human intentions and unlock their full potential, InstructGPT~\cite{instructgpt_2022} uses a small amount of supervised data to construct instruction-following data for fine-tuning LLMs and conducts reinforcement learning to train the model based on human preferences. This approach of instruction fine-tuning  has gained widespread adoption and following from both the academic and industrial communities~\cite{follow_instructgpt1,follow_instructgpt2,follow_instructgpt3,follow_instructgpt4,follow_instructgpt5,follow_instructgpt6}.

\begin{figure}[t!]
\begin{center}
     \scalebox{0.45}{           \includegraphics[width=1\textwidth]{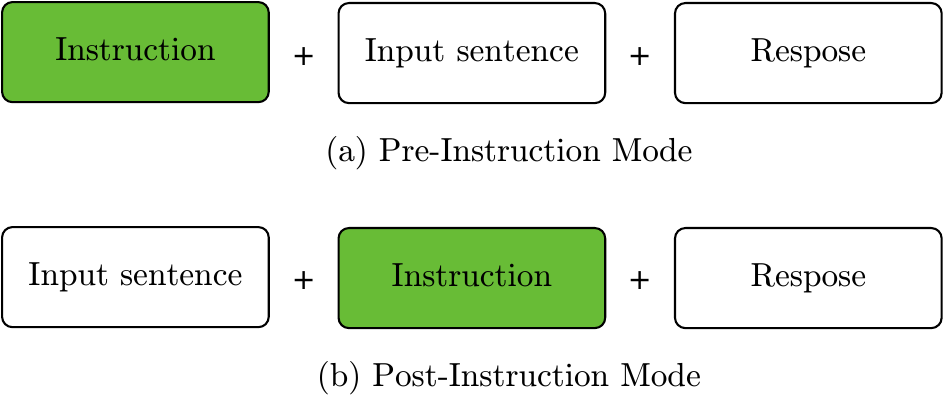}
      } 
      \caption{
       Example data in Pre-Instruction and Post-Instruction format. Different blocks represent textual data from different fields, while the `+' symbol signifies the concatenation operation for the textual data. For sequence generation tasks, the length of the input sentence is generally much larger than the length of the task instruction.
      } 
      \label{fig:post_pre_ins}  
 \end{center}
\end{figure}


Generally, the instruction-following data consists of three parts. Taking the machine translation task as an example, these parts include a specific task instruction ({\em e.g.,} "Please translate the following paragraph from English to French"), an input sentence (the English sentence to be translated), and the final response (the corresponding French translation). 
Since most large language models are based on the decoder-only structure of Transformer~\cite{transformer_2017,gpt-2,gpt3}, with a training objective  of next token predicting.
Generally, these three parts of the instruction-following data are sequentially concatenated into a long nature sentence as the input for language models. 
Given that the self-attention mechanism in Transformer decoders tends to focus more on nearby words {\em i.e.,} the locality of self-attention modeling~\cite{longformer_2020,reformer_2019,local_attn_2019}, there is a considerable risk of instruction forgetting when predicting responses for long input sentences. Such as performing long text summarization tasks, the input sentence may contain thousands of tokens.
Consequently, the model may be at risk of forgetting the initial task instruction when predicting responses, leading to the generation of responses that do not fully comply with the user's intent. In this paper, we refer to this issue as the \textit{instruction forgetting} issue.

To alleviate the above issue for LLMs during instruction fine-tuning, we first observe that the relative position of the input sentence and the task instruction is crucial. 
Therefore, we propose a simple and straightforward solution, namely, placing the task instruction at the end of the input sentence (referred to as `Post-Ins').
In this way, when the model predicts the final response, it effectively models the nearest preceding sequence, which is just the task instruction indicating what content should be generated next.
For comparison, we refer to the data format in existing studies where the task instruction is concatenated to the front of the input sentence as Post-Instruction (abbreviated as `Pre-Ins').

To verify whether Post-Ins improves the instruction-following ability of language models and alleviates the instruction forgetting issue on long sentences compared to Pre-Ins, we first analyze the conditional probability characteristics of the models under both data formats with the trinomial Bayes formula.
Through appropriate assumptions and formula derivations, we draw the following conclusions: (1) Pre-Ins tends to model a reverse conditional probability ({\em e.g.,} reverse translation probability), emphasizing the coverage of the input sentence while insufficiently modeling the task instruction. (2) Post-Ins is more inclined to model a conditional probability about the task instruction ({\em e.g.,} predicting the task instruction given inputs and outputs), emphasizing the modeling of task instruction-following ability. 

In addition to the theoretical analysis, we conduct extensive experiments based on two widely used large language models, LLaMA and BLOOMZ, with various parameter sizes ranging from 1.7 billion to 13 billion. 
We select two common sequence generation tasks as specific downstream tasks, namely,  machine translation and long text summarization.
The experimental results show that Post-Ins consistently outperforms Pre-Ins across various settings without using any additional supervised data. 
Furthermore, due to the superior modeling ability of task instruction, Post-Ins exhibits stronger task instruction generalization capabilities, resulting in significant performance gains in zero-shot translation tasks ({\em e.g.,} up to a 9.7 BLEU score improvement). 
Finally, human analysis results confirm that Post-Ins generates responses more faithful to user instructions, with a noticeable improvement on the issue of hallucination responses.


Our contribution can be summarized as follows:\footnote{Codes and data are at \url{https://github.com/Adaxry/Post-Instruction}}
\begin{itemize}
    \item We show that the position of task instruction is a key factor to conduct instruction fine-tuning with LLMs, and propose to relocate the task instruction after the input sequence ({\em i.e., } Post-Ins) could enhance the instruction-following ability of LLMs.
    \item Both our theoretical and experimental analyses demonstrate that Post-Ins pays larger attentions on the model's instruction-following capabilities, yielding consistent performance improvements across two common sequence generation tasks.
    

\end{itemize}

\section{Background}

\begin{figure*}[htbp]
  \centering
  \begin{subfigure}[b]{0.48\textwidth}
    \includegraphics[width=\textwidth]{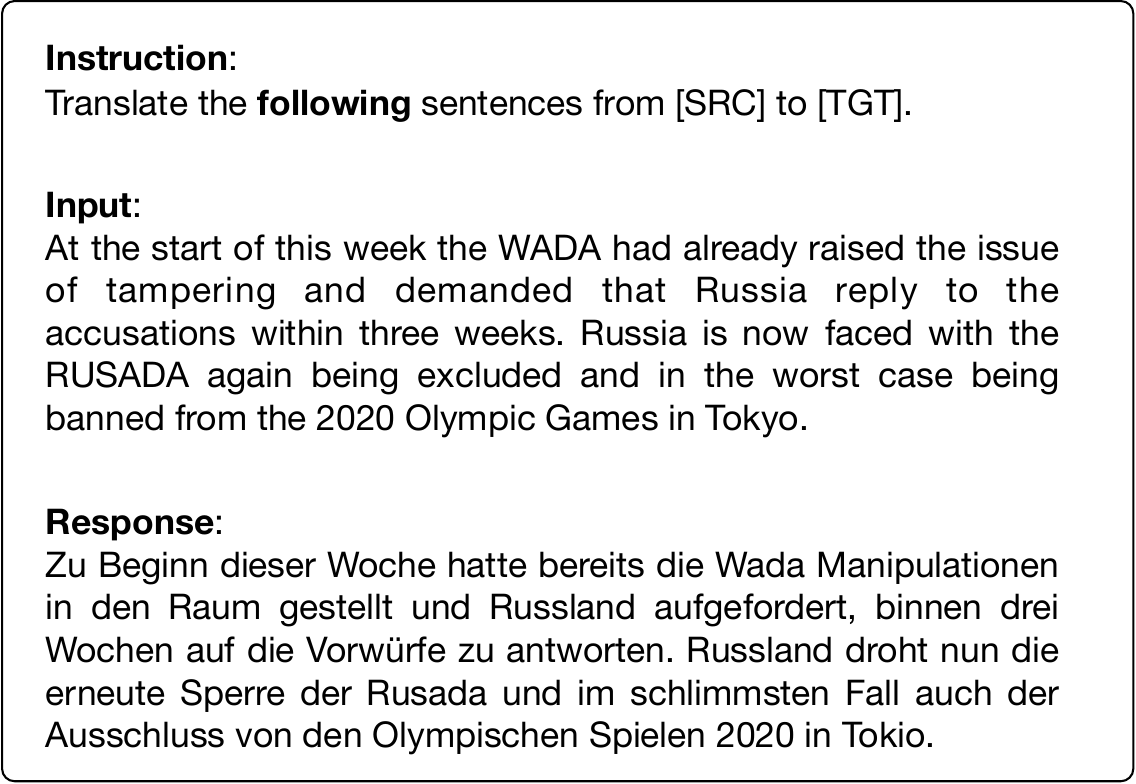}
    \caption{Pre-Ins}
    \label{fig:pre_ins_example}
  \end{subfigure}
  \hfill
  \begin{subfigure}[b]{0.48\textwidth}
    \includegraphics[width=\textwidth]{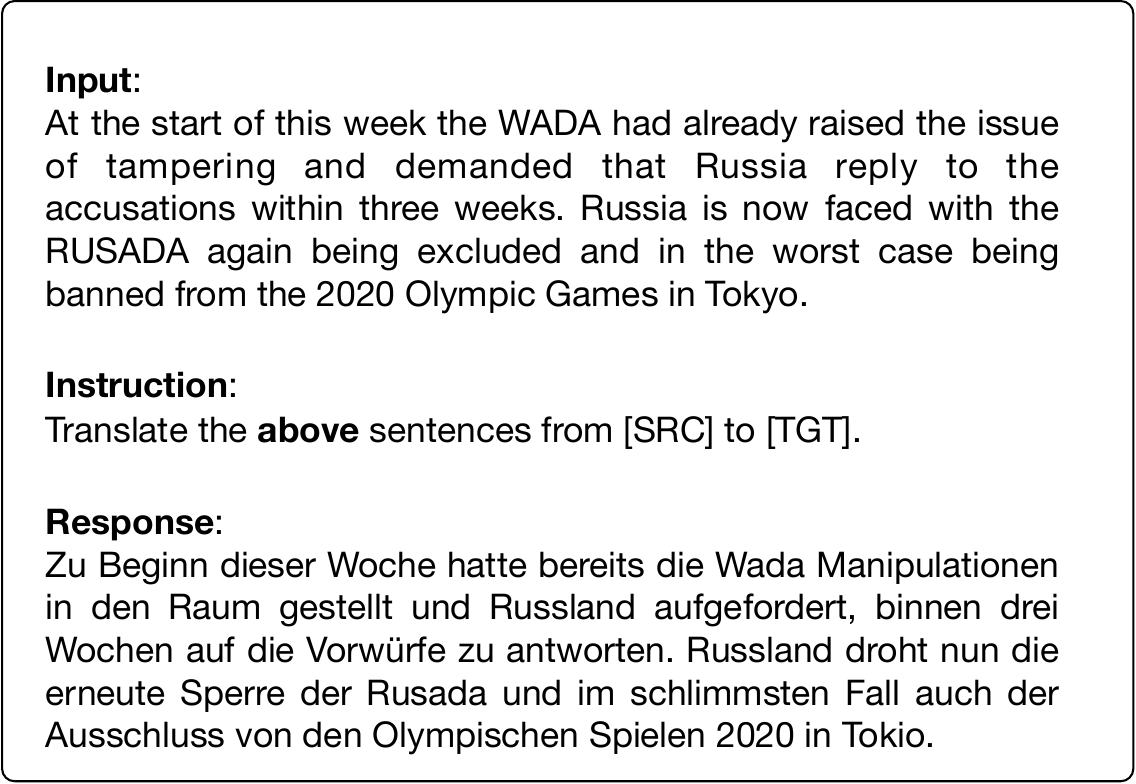}
    \caption{Post-Ins}
    \label{fig:post_ins_example}
  \end{subfigure}
  \caption{An example of Pre-Ins formatted data for the machine translation tasks. `[SRC]' and `[TGT]' refer to the source and target language, which are respectively English and German in this example.}
  \label{fig:post_vs_pre_example}
\end{figure*}

\paragraph{Instruction Fine-Tuning.}
InstructGPT~\cite{instructgpt_2022} is the first to unveil the immense potential of instruction learning, namely, a InstructGPT model with 1.3 billion parameters can outperform the 175B GPT-3, despite having 100x fewer parameters. Then Stanford release the Alpaca instruction-following datasets~\cite{alpaca}, which is constructed by the Self-instruction data generation pipline~\cite{self_instruct_2022}.
In the field of machine translation, Parrot~\cite{jiao2023parrot} builds contrastive and error-guided instructions to align the translation results of LLMs with humman preferences. 
Subsequently, ~\citep{zeng2023tim} further extend the error-guided instructions with token-level Direct Preference Optimization~\cite{dpo_2023}.
To better transfer the capabilities of sequence generation of LLMs, BayLing~\cite{bayling_2023} propose to conduct interactive translation task for instructing fine-tuning.
Although the aforementioned methods have made considerable progress, we argue that the Pre-Ins data format ultilized in existing studies face the potential risk of instruction forgetting issue, which is just we aims to address in this paper.



\section{Approach}

\subsection{Definition}
The standard instruction-following data format consists of three components: a specific task instruction $inst$, an input sentence $inp$, and the corresponding response $res$. Take the machine translation task as an example, $inst$ is a specific task instruction that directs the model to translate from the source language into the target language, while $inp$ and $res$ are respectively the source input sentence and target translation.
The $inst$, $inp$ and $res$ are then sequentially concatenated into a long sequence, which is then fed into the LLMs for training in a teacher-forcing mode\footnote{Following existing studies~\cite{alpaca,jiao2023parrot}, the cross-entropy loss is calculated merely on $res$, while $inp$ and $inst$ only participate in the forward encoding process.}.
We provide a specific training example in the Pre-ins format, as shown in Figure~\ref{fig:pre_ins_example}.

Considering the nature of sequence generation tasks, the input part ($ins$) often tends to be lengthy, such as translating an entire article or generating a summary of a paragraph. 
After applying the fine-grained tokenization, it can result in a long sequence of tokens for training. Generally, the mainstream LLMs are based the decoder-only architecture of Transformer, where the self-attention tends to pay larger attention on nearby tokens. 
Therefore, in the case of long sequences as mentioned above, there is a significant risk that the model may forget the frontmost task instruction in the Pre-Ins data format, yiedling responses that don not follow the task instruction.

\subsection{Preliminary Observations on Pre-Ins} 
\begin{figure}[t!]
\begin{center}
     \scalebox{0.45}{           \includegraphics[width=1\textwidth]{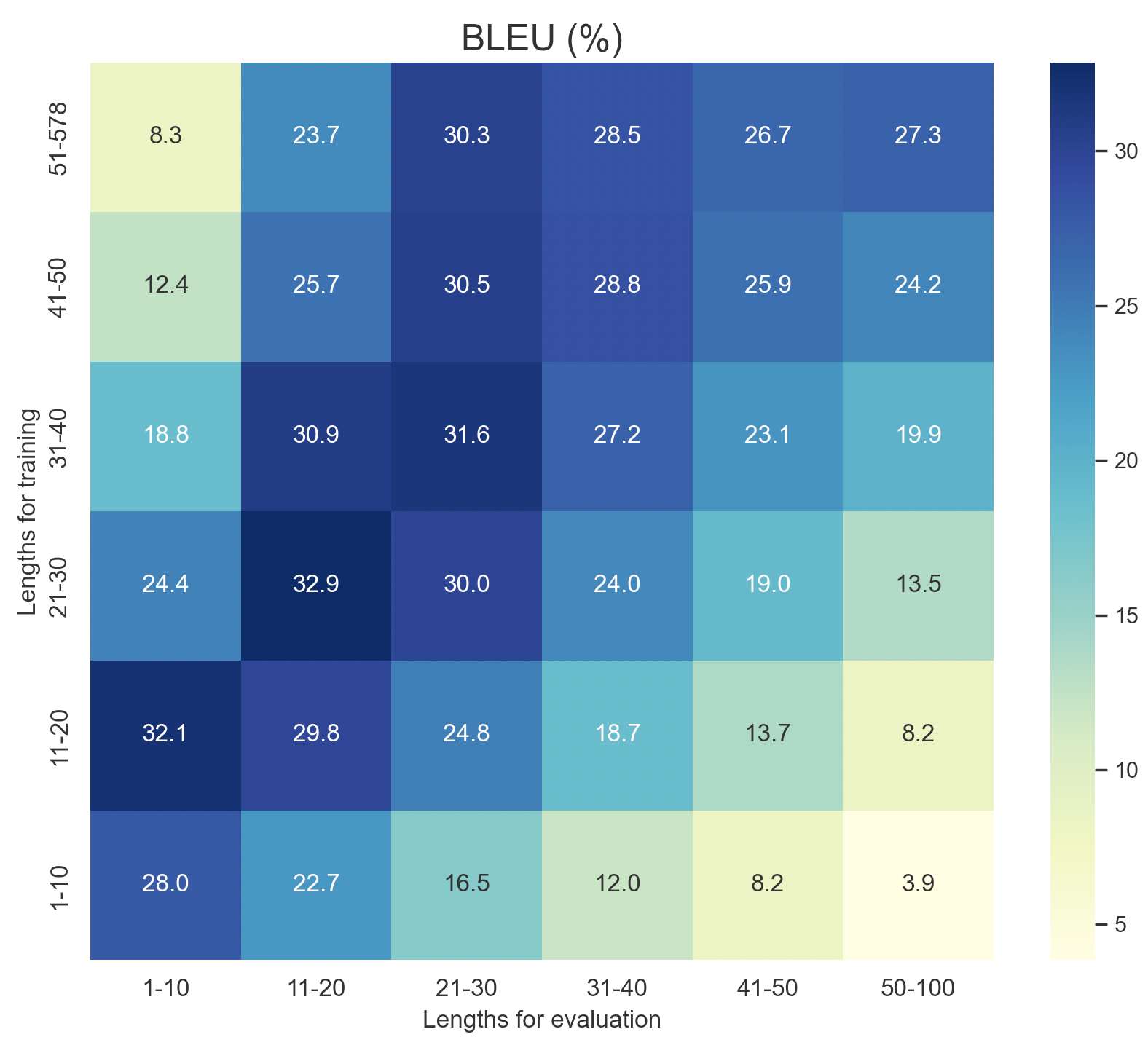}
      } 
      \caption{
        Performance of Pre-Ins over different length invertals in our preliminary experiments.
      } 
      \label{fig:len_interval}  
 \end{center}
\end{figure}

To verify whether the Pre-Ins data format  suffer from the above issue of instruction forgetting on long input sentences, we conduct preliminary experiments on the machine translation task (detailed experimental setups are at Section~\ref{sec:dataset}). We divide the training data into multiple groups based on the length range of the source text, ensuring that the total number of tokens in each group of training samples is approximately the same. Similarly, we also select corresponding test sets for different length ranges. Results on BLOOMZ are ploted in Figure~\ref{fig:len_interval}, we observe that the model tends to perform better on test sets with similar lengths to the training set, while it often struggles on test sets with different lengths (the diagonal line in Figure~\ref{fig:len_interval}).
Specifically, a model trained on short sentences may face difficulties in translating longer sentences due to limitations in the distribution of training data.
Interestingly, a model trained on longer sentences performs poorly on shorter sentence datasets (as shown in the top-left corner of the Figure~\ref{fig:len_interval}). Based on humman analysis, we discover a noticeable translation hallucination issue, where the model generates content that does not exist in the source text. 
These observations indicate that the existing Pre-Ins data format has limited ability to follow the instructions, especially when the input sentence $ins$ is long. Pre-Ins exhibits a risk of instruction forgetting, resulting in outputs that are not faithful to the user's intent.


\subsection{Post-Instruction}
To address the above issue of instruction forgetting, we propose a simple and straightforward solution, namely, relocating the task instruction $inst$ after the input sentence $inp$. As a result, the model can perceive the specific task instructions more closely when generating responses, regardless of the length of the input sentence. We refer this data format as Post-Instruction (Post-Ins), and provide a Post-Ins formatted example in Figure~\ref{fig:post_ins_example}.

Formally, the Post-Ins format of data encourages the LLMs to model the following conditional probability $p(res|inp, inst)$. Here, we can decompose the above formula using the trinomial Bayes' theorem as follows:

\begin{equation}
\small
\begin{aligned}
p(res|inp, inst) &= \frac{p(res) \cdot p(inp|res) \cdot p(inst|res, inp)}{p(inp) \cdot p(inst|inp) } \\
\end{aligned}
\end{equation}
where $p(inp|res)$ represents the probability of the input given the response. Considering that LLMs primarily encounter natural language during the pre-training phase, we can make the basic assumption that the input $inp$ and response $res$ are independent when no specific task instructions are given, namely, $p(inp|res) \approx p(inp)$. Therefore, we can further simplify the above formula as follows:
\begin{equation}
\small
\begin{aligned}
p(res|inp, inst) &= \frac{p(res) \cdot p(inp|res) \cdot p(inst|res, inp)}{p(inp) \cdot p(inst|inp) } \\
&\approx \frac{p(res) \cdot p(inst|res, inp)}{p(inst|inp) } \\
\end{aligned}
\end{equation}
Given that the task instruction $inst$ are not involved in the training loss, we can simply treat its predicted probability $p(inst|inp)$  as a constant, and get the following form:
\begin{equation}
\small
\begin{aligned}
p(res|inp, inst) = \underbrace{p(res)}_{fluency} \cdot \underbrace{p(inst|res, inp)}_{instruct} \cdot const
\end{aligned}
\label{equ:post_ins}
\end{equation}
where $p(res)$ denotes the modeling probability of the target response, which guarantees the fluency of the model in predicting the response.
On the other hand, $p(inst|res, inp)$  represents the probability of the model determining what task instruction is currently being executed given the input $inp$ and response $res$. This can ensure that the model has a strong perception of the requirements of the task instruction.

\subsection{Post-Instruction versus Pre-Instruction}
\label{sec:post_vs_ins}
As a comparison, we have also conducted a similar theoretical analysis for Pre-Ins, and ultimately obtain the following formula:
\begin{equation}
\small
\begin{aligned}
p(res|inst, inp) &= \frac{p(res) \cdot p(inst|res) \cdot p(inp|res, inst)}{p(inst) \cdot p(inp|inst) } \\
&\approx \frac{p(res) \cdot p(inp|res, inst)}{p(inp|inst) } \\
&= \underbrace{p(res)}_{fluency} \cdot \underbrace{p(inp|res, inst)}_{coverage} \cdot const
\end{aligned}
\end{equation}
Similar to Post-Ins, Pre-Ins also includes a component responsible for modeling the fluency of the response, denoted as $p(res)$. However, the key difference lies in the Pre-Ins emphasis on modeling the probability of the input given the instruction and response, namely, $p(inp|res, inst)$, which is similar to the modeling coverage in translation tasks~\cite{converage_2016}.
Such modeling approach may be suitable for a single task or a small number of training tasks, as the model can memorize these few task instructions through supervised fine-tuning. 

However, considering that LLMs inherently have strong fundamental capabilities that can naturally be applied to various sequence generation tasks, when modeling multiple sequence generation tasks simultaneously (such as multiple translation directions), Pre-Ins may suffer from instruction forgetting and produce low-quality responses that do not follow instructions due to the lack of task instruction modeling. 
In contrast, Post-Ins, with its preference for directly modeling task instructions as shown in Equation~(\ref{equ:post_ins}), can easily handle various sequence generation tasks and has good transferability for task instructions. We experimentally verify the stronger instruction transferability of Post-Ins compared to Pre-Ins in zero-shot translation tasks in section~\ref{para:zero_shot}. Furthermore, we analyz the modeling preferences of the two data formats from the perspective of attention distribution and observe that the conclusions are consistent with our previous theoretical analysis in section~\ref{sec:self_attn}.

\section{Experiments and Evaluations}

\subsection{Datasets}
\label{sec:dataset}
\paragraph{Alpaca.} 
The Alpaca dataset, released by Stanford~\cite{alpaca}, is widely used for instruction following tasks. It is constructed by the self-instruct data generation pipline~\cite{self_instruct_2022}, utilizing the text-davinci-003 model to generate high-quality instruction-following data. 
The data format follows the aforementioned Pre-Ins format, consisting of three parts: instruction, input, and output.
We adjust the positions of the instruction and input, yielding a Post-Ins formated Alpaca.
We apply this Post-Ins formatted Alpaca dataset to  experiments that are conducted in the Post-Ins format, while the other experiments are still conducted on the original Alpaca dataset.

\begin{table*}[t]
\begin{center}
\scalebox{0.85}{
\begin{tabular}{lrlcccccccc}
\toprule
\multirow{2}{*}{\textbf{Systems}} & \multirow{2}{*}{\textbf{\#Params}}  & \multirow{2}{*}{\textbf{Instruction}} & \multicolumn{4}{c}{\textbf{SacreBLEU}}  &  \multicolumn{4}{c}{\textbf{COMET22}} \\
~ & ~ & ~ & \multicolumn{2}{c}{De $\Longleftrightarrow$ En} & \multicolumn{2}{c}{Zh $\Longleftrightarrow$ En} & \multicolumn{2}{c}{De $\Longleftrightarrow$ En} & \multicolumn{2}{c}{Zh $\Longleftrightarrow$ En} \\

\midrule\midrule

\multicolumn{11}{c}{\textit{WMT22 Winners}} \\
WMT22 Winners & N/A & N/A & 33.70 & 38.40 & 33.50 & 54.30 & 85.46 & 88.09 & 81.12 & 87.84 \\
\midrule\midrule
\multicolumn{11}{c}{\textit{BLOOMZ-based}} \\
Parrot~\cite{jiao2023parrot} & 7.1B & Pre-Ins & 24.96 & 20.56 & 22.72  & 34.58 & 78.09 &  73.62 & 79.00 & 83.54  \\
TIM~\cite{zeng2023tim} & 7.1B & Pre-Ins & 24.31 & \textbf{20.63} & 23.42  & 37.20 &  77.65 & \textbf{ 74.16} & 79.50 & \textbf{84.89}  \\
\hdashline
\multirow{6}{*}{BLOOMZ} & 1.7B & Pre-Ins & 21.01 &	15.51 &	20.31 &	33.35   & 72.63 &	61.63 &	77.44 & 82.56  \\
 & 1.7B & Post-Ins & 20.99 &	16.68 &	20.15 &	34.02  & 73.76 & 63.64&	77.38&	82.97 \\
 
& 3.0B & Pre-Ins &  23.29 &	17.02 &	22.20 &	35.02 & 75.42 &	66.96	& 78.85 &	83.33   \\
& 3.0B & Post-Ins & 23.70 &	18.24 &	22.21 &	35.62  & 76.12	& 68.64 &	78.70 &	83.77 \\
& 7.1B & Pre-Ins &  24.37 &	19.77  & 22.98   & 36.64  & \textbf{78.45} &73.77&	\textbf{79.54} &	84.72  \\
& 7.1B & Post-Ins &  \textbf{25.46} & 19.79  & \textbf{23.65}  & \textbf{37.60}  & 77.79	& 72.79	 & 79.21 &	84.69 \\
\midrule \midrule
\multicolumn{11}{c}{\textit{LLaMA-based}} \\
Parrot~\cite{jiao2023parrot} & 7.0B & Pre-Ins & 27.38 & 26.14 & 20.23  & 30.33 & 82.47 &  81.67  & 75.90 & 80.34  \\
BayLing~\shortcite{bayling_2023} * & 7.0B & Pre-Ins & 28.16 & 25.66 & 20.31 & 38.19 &  83.19 &  82.18 & 77.48 & 84.43  \\ 
TIM~\cite{zeng2023tim} & 7.0B & Pre-Ins & 27.91 & 25.02 & 19.33  & 30.07 & 82.80 & \textbf{82.56} & 75.46 & {80.03} \\
BayLing~\shortcite{bayling_2023} * & 13.0B & Pre-Ins & 27.34 & 25.62 & 20.12 & 37.92 &  83.02 & 82.69 & 77.72 & 84.62  \\ 
TIM~\cite{zeng2023tim} & 13.0B & Pre-Ins & 29.03 & 26.71 & 20.27  & 32.14 & 83.48 & 83.31 & 76.64 & 81.30 \\
\hdashline
\multirow{4}{*}{LLaMA} & 7.0B & Pre-Ins & 29.98 & 25.23 &	17.68 &	23.83  & 82.63 & 81.27 & 72.90 & 75.70  \\
 & 7.0B & Post-Ins & \textbf{30.41} &	\textbf{26.50} &	\textbf{21.69} & \textbf{30.50}  & \textbf{83.62}	& 82.32	& \textbf{76.60} & \textbf{80.66}	 \\    
& 13.0B & Pre-Ins &  30.92 & 28.51  & 21.95  & 32.55 & 84.03 & 83.14 & 77.02 & 81.16 \\
& 13.0B & Post-Ins &  \textbf{31.25}& \textbf{28.70} & \textbf{22.37}  & \textbf{33.04} & \textbf{84.19} & 	\textbf{83.65} & \textbf{77.33 }& \textbf{82.16}  \\

\bottomrule
\end{tabular}
}
\caption{SacreBLEU and COMET22 score(\%) of different models with varying instruction modes on the WMT-2022 test sets. `De', `En' and `Zh' are the language code of `German', `English' and "Chinese", respectively. The \textbf{bolded} scores correspond to the best performance under the same or comparable settings for models with more than 7B parameters.
Results marked with `*' indicate that they are not directly comparable with other results due to the use of additional supervised data. 
}
\label{tab:wmt22}
\end{center}
\end{table*}

\paragraph{WMT Datasets.} 
The annual Conference on Machine Translation~\cite{wmt22, wmt21} provid high-quality human translations for evaluating the cutting edge machine translation systems.
In the experiments of this paper, we utilize the development sets from 2017 to 2020 as high-quality translation training data, following existing settings~\cite{jiao2023parrot,zeng2023tim}.
For the translation directions with multiple references, we duplicate the source side and then match them with the corresponding translations to form multiple translation sentence pairs. Finally, we obtrain a collection of 51k sentence pairs for instruction fine-tuning.
To facilitate comparison, we follow the settings of existing methods~\cite{jiao2023parrot,zeng2023tim} and fine-tune LLMs on data for three languages and four translation directions: Chinese-to-English, English-to-Chinese, German-to-English, and English-to-German.
The test sets for these four directions in WMT-2022 are used to evaluate translation performance, while the remaining directions, such as French-to-German or Russian-to-English, are used to evaluate the zero-shot performance of the models. 
Furthermore, considering the similarity in data distribution across the years in the WMT dataset~\cite{wmt20, wechat_wmt_2021}, we also conduct evaluation and validation on another test set, namely, the FLORES-200 benchmark.


\paragraph{Multidimensional Quality Metrics (MQM).} MQM dataset are based on the outputs of top systems from the WMT 2020 shared task, it provide error analysis of above translations annotated by professional translators. We follow the preprocessing scripts of existing studies and finally obtrain a same sized training set with 99k examples~\cite{jiao2023parrot,zeng2023tim}. In this paper, MQM is only used for translation task.

\paragraph{CNN/DailyMail.}
The popular CNN/DailyMail Dataset~\cite{cnndw_dataset_2017} is a collection of English-language news articles, comprising slightly over 300k unique articles authored by journalists from CNN and the Daily Mail.
The average sentence length of the source text of these data is about 665 words, or about one thousand tokens, which served as a widely used benchmark for the long text summarization~\cite{cnn1_2023,cnn2_2023,cnn3_2023}.
We follow the pre-processing and post-processing scrips of existing studies~\cite{prophetnet_2020}.
We use the  CNN/DailyMail dataset only for the text summarization task and conduct the evaluation on the standard test set with 11,490 samples.


\subsection{Evaluation}
\paragraph{Inference Settings.} For all tasks, we set the batch size to 1 during inference to avoid the effect of padding side ({\em e.g,} BLOOMZ applies left-padding mode, while LLaMA uses right-padding mode when batching the input data). 
We set the temperature coefficient to 0.1 in order to encourage the model to output more accurate rather than diverse responses for these conditional sequence generation tasks.
As for the decoding strategies, we apply the beam search for all task, and set beam size to 4 for machine translation. While for the text summarization task, we have to decrease the beam size to 2, as encoding the long input sentences will consume a large portion of GPU memory.

\paragraph{Metrics}
For the machine translation task, we use SacreBLEU\footnote{\url{ https://github.com/mjpost/sacrebleu}} to calculate BLEU scores. Given the limitations of N-gram based metrics  to measure semantic similarity, we also calculate the popular neural-based metric, namely, COMET22\footnote{\url{https://github.com/Unbabel/COMET}}.
For the text summarization task, we report the F1 scores of ROUGE-1, ROUGE-2, and ROUGE-L following existing studies~\cite{cnn1_2023,prophetnet_2020}.

\begin{table*}[t]
\begin{center}
\scalebox{0.85}{
\begin{tabular}{lrlcccccccc}
\toprule
\multirow{2}{*}{\textbf{Systems}} & \multirow{2}{*}{\textbf{\#Params}}  & \multirow{2}{*}{\textbf{Instruction}} & \multicolumn{4}{c}{\textbf{SacreBLEU}}  &  \multicolumn{4}{c}{\textbf{COMET22}} \\
~ & ~ & ~ & \multicolumn{2}{c}{De $\Longleftrightarrow$ En} & \multicolumn{2}{c}{Zh $\Longleftrightarrow$ En} & \multicolumn{2}{c}{De $\Longleftrightarrow$ En} & \multicolumn{2}{c}{Zh $\Longleftrightarrow$ En} \\

\midrule 
TIM~\cite{zeng2023tim} & 7.0B & Pre-Ins & 39.15 & 29.31 & \textbf{22.30} & 28.43  & 88.19 & 85.05 & \textbf{83.32} & 80.55 \\
\hdashline
\multirow{4}{*}{LLaMA} & 7.0B & Pre-Ins & 38.86  & 29.51 &  18.10 & 21.69  & 88.05 & 84.57 &80.69 & 75.07  \\
 & 7.0B & Post-Ins & \textbf{41.12} & \textbf{31.27} & 21.80 & \textbf{28.89} &  \textbf{88.63} & \textbf{85.53} & \textbf{83.57} & \textbf{81.23}  \\ 
& 13.0B & Pre-Ins & \textbf{41.78} &  \textbf{33.62} & 22.21 &  30.74  & 88.91 & \textbf{86.36} &  84.26 &  82.50  \\
& 13.0B & Post-Ins & 41.46 &  33.12 &  \textbf{22.62} & \textbf{31.37} & 88.91 & 86.23  & \textbf{84.38} & \textbf{82.83} \\

\bottomrule
\end{tabular}
}
\caption{SacreBLEU and COMET22 score(\%) of different models with varying instruction modes on the FLORES-200 test sets. The \textbf{bolded} scores correspond to the best performance under the same or comparable settings.}
\label{tab:flores}
\end{center}
\end{table*}

\begin{table*}[t]
\begin{center}
\scalebox{0.81}{
\begin{tabular}{lrlrrrrrrrrrrrrr}
\toprule
\multirow{2}{*}{\textbf{Systems}} & \multirow{2}{*}{\textbf{\#Para.}}  & \multirow{2}{*}{\textbf{Ins.}} & \multicolumn{13}{c}{\textbf{SacreBLEU}} \\
~ & ~ & ~ & \multicolumn{2}{c}{Cs $\Leftrightarrow$ En} & \multicolumn{2}{c}{De $\Leftrightarrow$ Fr} & \multicolumn{2}{c}{Ja $\Leftrightarrow$ En} & \multicolumn{2}{c}{Uk $\Leftrightarrow$ En} & \multicolumn{2}{c}{Ru $\Leftrightarrow$ En} & \multicolumn{2}{c}{Liv $\Leftrightarrow$ En} & Average\\
\midrule
\midrule
\multirow{2}{*}{BLOOMZ} & 7.0B & Pre-Ins & \textbf{6.0} & 4.3 & 15.6 & \textbf{23.0} & 11.0 & 2.5 & \textbf{11.0} & 1.9 & \textbf{21.7} & 5.8 & \textbf{3.0} & 3.5 & 9.1 \\
 & 7.0B & Post-Ins &  5.6 & \textbf{4.5}  &  \textbf{24.4} & 22.7 &  11.0 & \textbf{2.6} & 10.2 & \textbf{2.0} & 21.3 & \textbf{6.1} & 2.9 & \textbf{4.2} & \textbf{9.8}  \\ 
\hdashline
\multirow{4}{*}{LLaMA} & 7.0B & Pre-Ins & 36.8 & 13.7 & 3.0  & 3.4 & \textbf{12.2} & 4.8 & 33.9 & 4.6 & \textbf{34.8} & 16.8 & \textbf{5.9} & 2.6 & 14.3  \\
 & 7.0B & Post-Ins & \textbf{36.8} & \textbf{17.4}  & \textbf{3.2} & \textbf{8.8} & 12.1  & \textbf{7.3} & \textbf{34.6}  & \textbf{11.7} & 34.7 & \textbf{18.9} &  5.0 & \textbf{3.3} & \textbf{16.2}  \\ 
& 13.0B & Pre-Ins & \textbf{39.5} & 19.7  & 4.9 & 27.5  & \textbf{13.9} & 3.4 & \textbf{36.8} & 17.2 & \textbf{37.6} & \textbf{21.1} & 5.5 & 2.9 & 19.1 \\
& 13.0B & Post-Ins & 36.8 & \textbf{19.7} & \textbf{14.6} & \textbf{33.3} & 13.6 & \textbf{6.0} & 35.6 & \textbf{17.6} & 36.2 & 20.8 & \textbf{5.6} & \textbf{3.0} & \textbf{20.2} \\

\bottomrule
\end{tabular}
}
\caption{SacreBLEU score(\%) of different models with varying instruction modes on the WMT-2022 zero-shot test sets. The \textbf{bolded} scores correspond to the best performance under the same or comparable settings. `Para.' is short for `Parameters' and `Ins.' stands for the data format for the instruction-following data. `CS', `Uk', `Ja', `Ru' and 'Liv' are the language code for `Czech', `Ukrainian', `Japanese', `Russian' and `Livonian', respectively.}
\label{tab:zero_shot}
\end{center}
\end{table*}

\section{Main Results and Analysis}
In this section, we first list the detailed experimental results of both the machine translations and text summarization tasks in Section~\ref{sec:mt} and Section~\ref{sec:sum}. 
Subsequently, we show the analysis of the  distributions of self-attention in Section~\ref{sec:self_attn}, and human evaluation results in Section~\ref{sec:humman_analysis}.

\subsection{Results of Machine Translation}
\label{sec:mt}
\paragraph{Supervised Translation.}
Table \ref{tab:wmt22} presents a summary of experimental results on WMT22. Our proposed method, Post-Ins, consistently outperforms Pre-Ins in most settings of BLOOMZ. The maximum improvement in the BLEU score is observed to be +1.22, specifically in the En$\Rightarrow$De direction of the BLOOMZ-3B. Additionally, COMET22 scores reach up to +2.01 improvement in the Ee$\Rightarrow$De translation using the 1.7B model. Our method proves to be even more effective in the LLaMA model, outperforming Pre-Ins in all settings. Particularly, LLaMA-7B achieves a remarkable increase of +6.67 BLEU and +4.96 in COMET22 in En$\Rightarrow$Zh translation. 
It is worth noting that Post-Ins also surpasses existing translation approaches that based on BLOOMZ and LLaMA.
Table \ref{tab:flores} showcases the performance of our method on the Flores-200 test set. Our approach outperforms Pre-Ins in 13 out of 16 settings, with maximum improvements reaching +7.20 BLEU and +6.16 COMET22 score in En$\Rightarrow$Zh.

Specially for the LLaMA, the improvement in En$\Leftrightarrow$Zh translation is significantly larger than that in EN$\Leftrightarrow$DE translation. 
We hypothesize that this is due to the fact that the LLaMA vocabulary splits Chinese into individual characters, resulting in longer sentence lengths and an increase in the number of tokens, which align with 
the discussion in Section~\ref{sec:post_vs_ins}, namely, Post-Ins are naturally better equipped to handle generation for long sentence. 

\begin{table}[t]
\begin{center}
\scalebox{0.9}{
\begin{tabular}{llccc}
\toprule
{\textbf{\#Params}}  & {\textbf{Instruction}} & \textbf{RG-1} & \textbf{RG-2}  & \textbf{RG-L}  \\
\midrule \midrule
\multicolumn{5}{c}{\textit{BLOOMZ-based}} \\
 3.0B & Pre-Ins & 35.41 & 16.33 & 25.81  \\
 3.0B & Post-Ins & \textbf{38.90} & 
 \textbf{17.84} & \textbf{27.67}  \\
 \hdashline
 7.0B & Pre-Ins &  37.54 & 17.04 & 26.90  \\
 7.0B & Post-Ins & \textbf{38.61} & \textbf{17.64} & \textbf{27.49}   \\
\midrule \midrule
\multicolumn{5}{c}{\textit{LLaMA-based}} \\
7.0B & Pre-Ins & 37.55 & 17.17 & 26.30   \\
7.0B & Post-Ins & \textbf{38.11} & \textbf{17.66} & \textbf{26.88}  \\
\bottomrule
\end{tabular}
}
\caption{F1 scores of ROUGE-1 / ROUGE-2 / ROUGE-L on the test set of CNN/DailyMail. `RG' is an abbreviation for `ROUGE'.
The \textbf{bolded} scores correspond to the best performance under the same or comparable settings.}
\label{tab:cnndm}
\end{center}
\end{table}

\begin{figure*}[htbp]
  \centering
  \begin{subfigure}[b]{0.48\textwidth}
    \includegraphics[width=\textwidth]{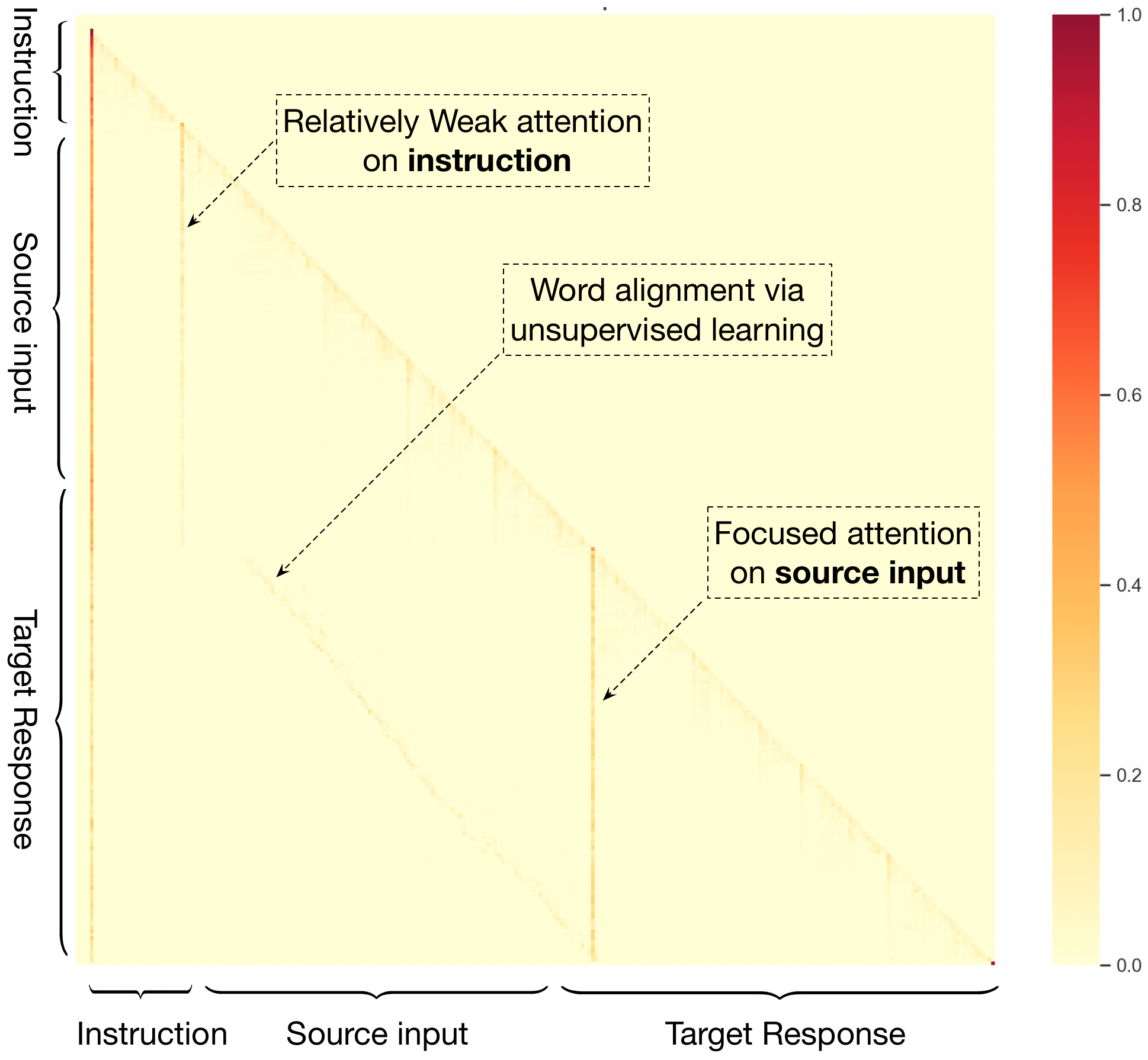}
    \caption{Attention heatmap of Pre-Ins.}
    \label{fig:post_heatmap}
  \end{subfigure}
  \hfill
  \begin{subfigure}[b]{0.48\textwidth}
    \includegraphics[width=\textwidth]{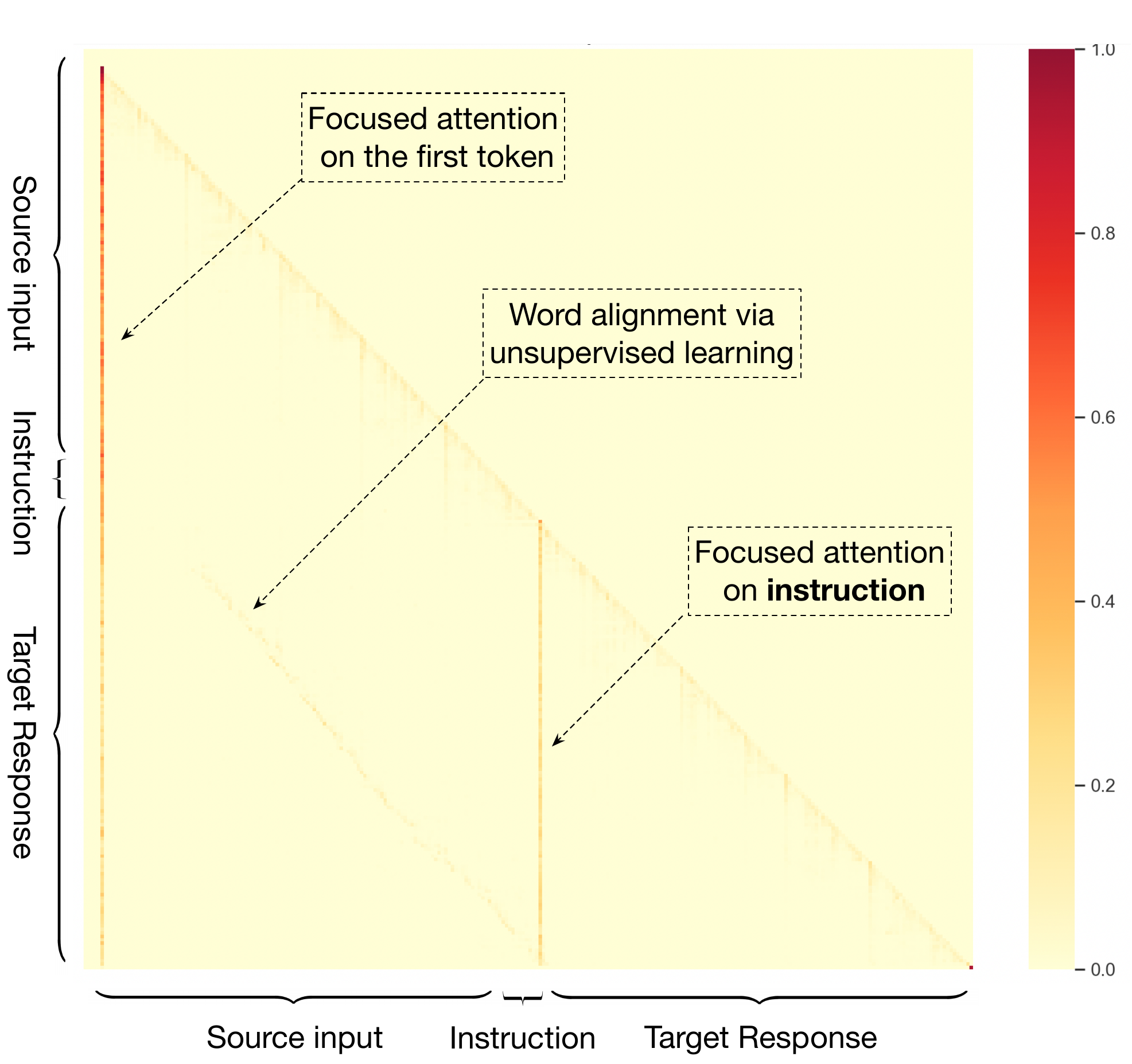}
    \caption{Attention heatmap of Post-Ins.}
    \label{fig:pre_heatmap}
  \end{subfigure}
  \caption{The visualization analysis of self-attention for the instruction fine-tuned BLOOMZ-7.1B model, where thicker lines indicate higher attention for the corresponding positions, while thinner lines indicate lower attention.}
  \label{fig:post_vs_pre_heatmap}
\end{figure*}

\begin{figure*}[t!]
\begin{center}
     \scalebox{1.}{           \includegraphics[width=1\textwidth]{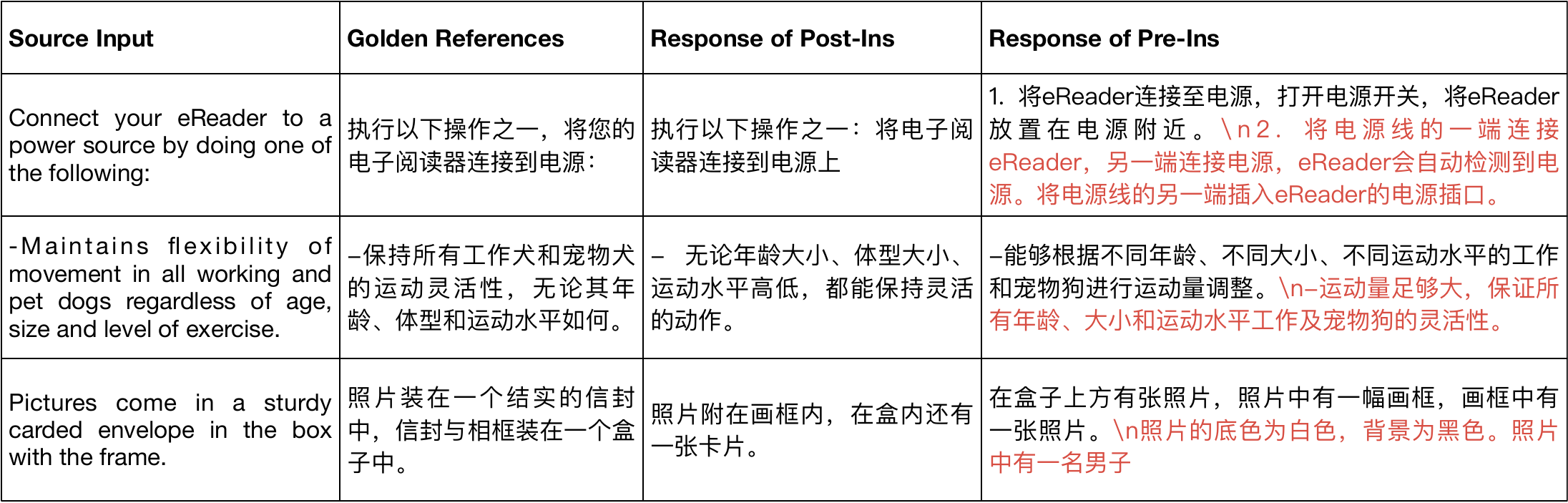}
      } 
      \caption{
        Case studies on Chinese-to-English translation tasks, where the highlighted red texts indicate the model deviates from the translation instruction, generating content not present in the source text.
      } 
      \label{fig:show_case}  
 \end{center}
\end{figure*}

\paragraph{Zero-Shot Translation.}
\label{para:zero_shot}
Furthermore, we observe significant improvements in zero-shot translation under the Post-Ins mode. 
Table \ref{tab:zero_shot} reports the results of different instruction modes on the WMT22 zero-shot test set. In terms of BLOOMZ, there is an impressive increase of +8.8 in De$\Rightarrow$Fr translation, with an average improvement of +0.7 BLEU.
For LLaMA-7B, an average improvement of +1.9 BLEU is achieved, while LLaMA-13B exhibit an average improvement of +1.1 BLEU. Notably, LLaMA-13B showcase the highest improvement of +9.7 BLEU in De$\Rightarrow$Fr translation. 
Overall, the consistent improvements of Post-Ins over Pre-Ins indicate that Post-Ins exhibits stronger instruction generalization capabilities, being able to generate responses effectively even for task instructions it has never encountered during fine-tuning.

\subsection{Results of Text Summarization}
\label{sec:sum}
To further validate whether Post-Ins can effectively alleviate the issue of instruction forgetting, we conduct experiments on tasks the long text summarization task, where the average lengths of input tokens is over 1,000.
Table \ref{tab:cnndm} presents the experimental results for the text summarization task in the CNN/DailyMail, where we report the F1 scores of ROUGE-1, ROUGE-2, and ROUGE-L. It is evident that all models achieved significant performance improvements up to +3.49 in BLOOMZ-3B when utilizing the Post-Ins approach. 
The effectiveness of Post-Ins on the task of long sequence generation is further demonstrated by the superiority of its effect on the text summarization task.

\subsection{Distributions of Self-attention}
\label{sec:self_attn}
Given that the distribution of self-attention can explain the behavior of the transformer model to some extent~\cite{self_attn_2021,self_attn1,self_attn2,self_attn3}, we plot the heatmap of self-attention for models trained with Pre-Ins and Post-Ins in Figure \ref{fig:post_vs_pre_heatmap}.
We take BLOOMZ-7.1B as the base model and conduct forward propagation on the training samples of machine translation to obtrain the attention scores.
To mitigate the impact of fluctuations of multi-head attention and various layers, we average the scores of all heads over different layers to obtain the final score.
We reach the following observations:
\begin{itemize}
\item A higher distribution of attention scores is observed at the beginning of sentences and along the diagonal positions of the attention matrix, which is consistent with existing conclusions~\cite{u_atten_2023}.

\item As shown in the lower right part of the Figure~\ref{fig:pre_heatmap}, Post-Ins pay more attentions on the specific task instruction when generating the response, while Pre-Ins mainly foucs on the source input and pay weak attention on instruction as shown in the upper left part of the Figure~\ref{fig:post_heatmap}.

\item After instruction fine-tuning on the machine translation data, models learn latent word alignment information on both data format. Namely, models tend to allocate more attention to the aligned parts of the source when generating responses word by word, which is similar to the conclusions of the traditional encoder-decoder structure in the field of machine translation~\cite{seq2seq,unsupervised_nmt}.
\end{itemize}
In summary, through the visualization analysis of the self-attention heatmap, we observe that Post-Ins naturally tends to model task instructions, while Pre-Ins relatively emphasizes task instructions to a weaker extent. This finding is consistent with the theoretical analysis and conclusions presented earlier in Section~\ref{sec:post_vs_ins}.

\subsection{Human Analysis}
\label{sec:humman_analysis}
We employ two linguistics professionals to evaluate the English-to-Chinese translation task. Specifically, the annotators are requested to judge whether the model's output faithfully follow to the translation instructions and the source input. 
If translation hallucinations occurred, {\em i.e.,} the response contain content that is not present in the source sentence, or if the translation task is not effectively executed, the label was marked as `0'; otherwise, it is marked as `1'. The manual annotation results on 1,000 samples show that the hallucination rate for Pre-Ins is 4.8\%, while 1.7\% for Post-Ins. We also provid several examples in the Figure~\ref{fig:show_case}. 
In summary, Post-Ins can enhance the model's instruction-following capability and effectively reduce the proportion of prediction hallucinations.

\section{Conclusion}
In conclusion, this paper highlights the importance of task instruction positioning in the instruction fine-tuning process of large language models (LLMs) for conditional sequence generation tasks. We propose a simple yet effective method, Post-Ins, which relocates the task instruction after the input sequence to enhance the instruction-following ability of LLMs. Our theoretical analysis and experimental results demonstrate that Post-Ins effectively shifts the model's learning focus, leading to improved performance across various model scales and different sequence generation tasks, such as machine translation and long text summarization. Notably, our method significantly boosts zero-shot performance without incurring additional data or annotation costs. These findings suggest that the proposed Post-Ins approach is a promising direction for further research and practical applications in the field of natural language processing and large-scale language models.

\bibliography{anthology,custom}
\bibliographystyle{acl_natbib}


\end{document}